\documentclass{article} 
\usepackage{iclr2015,times}
\usepackage{hyperref}
\usepackage{url}
\usepackage{amssymb}
\usepackage{amsthm}
\usepackage{graphicx}

\newcommand{\x}{\mathbf x}
\newcommand{\y}{\mathbf y}
\newcommand{\W}{\mathbf W}

\newcommand{\h}{\mathbf h}

\newcommand{\bias}{\mathbf b}
\newcommand{\Reals}{\mathbb{R}}
\title{Generative class-conditional denoising autoencoders}

\author{
Jan Rudy \& Graham Taylor\\
School of Engineering\\
University of Guelph\\
Guelph, Ontario, Canada \\
\texttt{\{jrudy,gwtaylor\}@uoguelph.ca} \\
}

\theoremstyle{plain}
\newtheorem{thm}{Theorem}

\iclrfinalcopy

\begin{document}

\maketitle

\begin{abstract}
  Recent work by \citet{bengio2013generalized} proposes
  a sampling procedure for denoising autoencoders which involves learning the transition
  operator of a Markov chain. 
  The transition operator is typically unimodal, which limits its capacity
  to model complex data.
  In order to perform efficient sampling from conditional distributions,
  we extend this work, both theoretically and algorithmically, to gated autoencoders \citep{memisevic2013learning},
  The proposed model is able to generate convincing class-conditional samples when
  trained on both the MNIST and TFD datasets.
\end{abstract}

\section{Introduction}
In the field of deep neural networks, purely supervised models trained
on massive labeled datasets have garnered much attention over the last few years \citep{dahl2010phone,deng2010binary,krizhevsky2012imagenet,goodfellow2014multi,szegedy2014going}.
However, recent work 
has rekindled interest in generative models of data \citep{bengio2013generalized,bengio2013deep,kingma2014auto,goodfellow2014generative}.
The recently proposed sampling procedure for denoising autoencoders \citep{bengio2013generalized} and 
their generalization to Generative Stochastic Networks \citep{bengio2013deep}
presents a novel training procedure which, instead of attempting
to maximize the likelihood of the data under the model, amounts to learning
the transition operator of a Markov chain \citep{bengio2013generalized}.

Although these models have shown both theoretically and empirically to
have the capacity to model the underlying data generating
distribution, the unimodal transition operator learned in
\citep{bengio2013generalized} and \citep{bengio2013deep} limits the
types of distributions that can be modeled successfully. One way to
address this issue is by adopting an alternative generative model such
as the Neural Autoregressive Density Estimator (NADE) as the output
distribution of the transition operator \citep{ozair2013multimodal}.

In this paper, we propose a alternate approach. Our main motivation is
that when labeled data is available, we can use the label information in order to carve up the
landscape of the data distribution. Although our model is generative, 
this work shares architectural similarities with discriminative models
such as the Hybrid Discriminative Restricted Boltzmann Machine \citep{larochelle2008classification},
the gated softmax \citep{memisevic2010gated}, and discriminative
fine-tuning of class-specific autoencoders \citep{kamyshanska2013autoencoder}.

This work begins by presenting an overview of related work, 
including a treatment of autoencoders, denoising autoencoders,
autoencoders as generative models and gated autoencoders.
Next, we propose a class-conditional gated autoencoder
along with training and sampling procedures based on the work
of \citet{bengio2013generalized}. Finally,
we present experimental results of our model on two image-based datasets.

\section{Related work}
\subsection{Autoencoders}
An autoencoder is a feed-forward neural network which aims
to minimize the reconstruction error of an input data vector via 
a latent representation. This can be interpreted
as the composition of two learned functions, the encoder function
$f$ and the decoder function $g$.
The encoder function $f$ is a mapping from input space onto the
representation space. Formally, 
given input vector 
$\x \in \Reals^{n_X}$, input weights $\W \in \Reals^{n_H \times n_X}$ and
hidden biases $\bias_{h} \in \Reals^{n_H}$, 
\begin{equation}
  \label{eq:ae_enc}
  f(\x) = s_H(\W \x + \bias)
\end{equation}

where $n_H$ is the dimension of the hidden representation and $s_H$
is an activation function. The activation is a non-linear function,
often from the sigmoidal family 
(e.g.~the logistic or hyperbolic tangent functions) or a piecewise linear function (e.g. rectified linear). The decoder then
projects this representation $\h = f(\x)$ back onto the input space, 

\begin{equation}
  \label{eq:ae_dec}
  \hat \x = g(\h) = s_O(\W' \h + \bias')
\end{equation}

where $\hat \x$ is the reconstruction of the input vector,
$s_O$ is the output activation function, $\W' \in \Reals^{n_X \times n_H}$
are the output weights and $\bias' \in \Reals^{n_X}$ are the output
biases. In order to restrict the number of free parameters of the model,
the  input and output weight matrices are often `tied' such
that $\W' = \W^{T}$.

The model parameters (i.e.~weights and biases) are updated via
a gradient-based optimization in order to minimize a loss function $L(\x)$
based on reconstruction error. The choice of loss function depends
on the data domain. When dimensions of $\x$ are real-valued, 
a typical choice of $L(\x)$ is the squared error, i.e. $L(\x) = \sum_{i=1}^{n_X}(x_i - \hat x_i)^2$.
When $\x$ is binary, a more appropriate loss function is the cross-entropy
loss, i.e. $L(\x) = \sum_{i=1}^{n_X} x_i \log \hat x_i + (1-x_i) \log (1 - \hat x_i)$.

\subsection{Denoising Autoencoders}
When the dimension of the hidden representation $n_H$ is smaller than
the dimension of the data space $n_X$, the learning procedure encourages
the model to learn the underlying structure of the data. The data
representation can exploit structure in order to compress the 
data to fewer dimensions than the original space. As such, each dimension
of the representation space is interpretable as a useful feature of the data.

However,
when $n_H \geq n_X$, the autoencoder can achieve perfect reconstruction
without learning useful features in the data by simply learning
the identity function. In this so-called ``overcomplete'' setup,
regularization is essential. Among the various kinds of regularized
autoencoders,
the denoising autoencoder (DAE) \citep{vincent2008extracting} is among
the most popular and well-understood. 
Instead of reconstructing the data from the actual input vector $\x$,
the DAE attempts to reconstruct the original input from an encoding of
a corrupted version, $f(\tilde \x)$. This effectively
prohibits the model from learning a trivial solution while learning
robust features of the data.

The corruption procedure is defined as a sample from the conditional
distribution $\tilde \x \sim C(\tilde \x | \x)$. Here, $\tilde \x$ is 
a noisy version of $\x$ where the type of noise is defined by the distribution
$C(\tilde \x | \x)$. Typical choices of noise are

\begin{enumerate}
  \item 
    Gaussian noise - adding $\epsilon_i \sim N(0, \sigma)$
    to each dimension.
    
  \item 
    Masking noise - setting $x_i = 0$ with probability $\rho$,
    
  \item
    Salt-and-pepper noise - similar to masking noise, but 
    corrupting $x_i$ with probability $\rho$ and each
    corrupted dimension is set to $x_i = 0$ or $x_i = 1$ with
    probability 0.5.
\end{enumerate}

Apart from preventing the model to learn a trivial representation
of the input by acting as a form or regularization, the DAE can be interpreted as a
means of learning the manifold of the underlying data generating
distribution \citep{vincent2008extracting}. Under the assumption
that the data lies along a low dimensional manifold in the input
space, the corruption procedure is a means of moving the training data
away from this manifold.  In order to successfully
denoise the corrupted input, the model must learn to project
the corrupted input back onto the manifold. Under this
interpretation the hidden representation can be interpreted as
a coordinate system of manifolds \citep{vincent2008extracting}.

\subsection{Denoising autoencoders as generative models}
Although DAEs are useful as a means of pre-training
discriminative models, especially when stacked to form deep models \citep{vincent2010stacked},
recent work by \citet{bengio2013generalized} has shown that
DAEs and their variants locally characterize the data generating
density. This provides an important link between DAEs and
probabilistic generative models.

We define observed data $\x$ such that $\x \sim \mathcal{P}(\x)$ where $\mathcal{P}(\x)$ is the
true data distribution and we define $C(\tilde \x | \x)$ as the conditional
distribution of the corruption process. 
When such models are trained using a loss function that can be
interpreted as log-likelihood, by predicting $\x$ given $\tilde \x$, the
model is learning the conditional distribution $\mathcal{P}_{\theta}(\x | \tilde \x)$
(where $\theta$ represents the parameters of the model).

In order to generate samples from the model, one simply forms a Markov
chain which alternately samples from learned distribution and the corruption distribution \citep{bengio2013generalized}.
Where $\x_t$ is the state of the Markov chain at time $t$, then
$\x_{t} \sim P_{\theta}(\x | \tilde \x_{t-1})$ and
$\tilde \x_{t} \sim C(\tilde \x | \x_{t})$. In other words, samples can
be generated by alternating between the corruption process and the 
denoising function learned by the autoencoder.

Notice that this is not a true Gibbs sampling procedure, as $(\x_{t}, \tilde \x_{t-1})$
may not share the same asymptotic distribution as $(\x_{t}, \tilde \x_{t})$. Regardless,
theoretical results indicate that under some conditions elaborated on in Sec.~\ref{sec:classcond}, the
asymptotic distribution of the generated samples converges to the
data-generating distribution \citep{bengio2013generalized}.

Although the above procedure does produce convincing samples,
it will also generate spurious examples that lie far from
any of the training data. This is because the training procedure does not sufficiently explore the
input space.
Under the manifold interpretation described above, the corruption
procedure defines a region around each example that is 
explored during training, the size of which is determined by the
amount of the corruption (e.g.~$\sigma$ in the case of Gaussian
noise and $\rho$ for masking or salt and pepper noise). This can
leave much of the input space unexplored, allowing the model
to place appreciable amounts of probability mass (i.e. spurious modes) in regions
of the space that lie far from any training example.

One solution involves using large or increasing amounts of noise
during training, however this results in a naive search of the space.
A more efficient procedure called ``walkback training'' is described in \citep{bengio2013generalized} and
bears resemblance to contrastive divergence training of restricted Boltzman machines.
Instead of defining the loss as the reconstruction cost of a single step
of corruption and reconstruction chain, walkback training defines a
series of $k$ reconstructions via a random walk originating at the training example.
Each reconstruction is corrupted and subsequently reconstructed,
where the final cost is defined as the sum of the reconstruction
costs of each intermediate reconstruction.
Since the training procedure mimics that of the sampling procedure,
walkback training is a means of seeking out these spurious modes and
redistributing their probability mass back towards the manifold.

\subsection{Gated autoencoders}

\begin{figure}
  \begin{center}
    \includegraphics[scale=0.7]{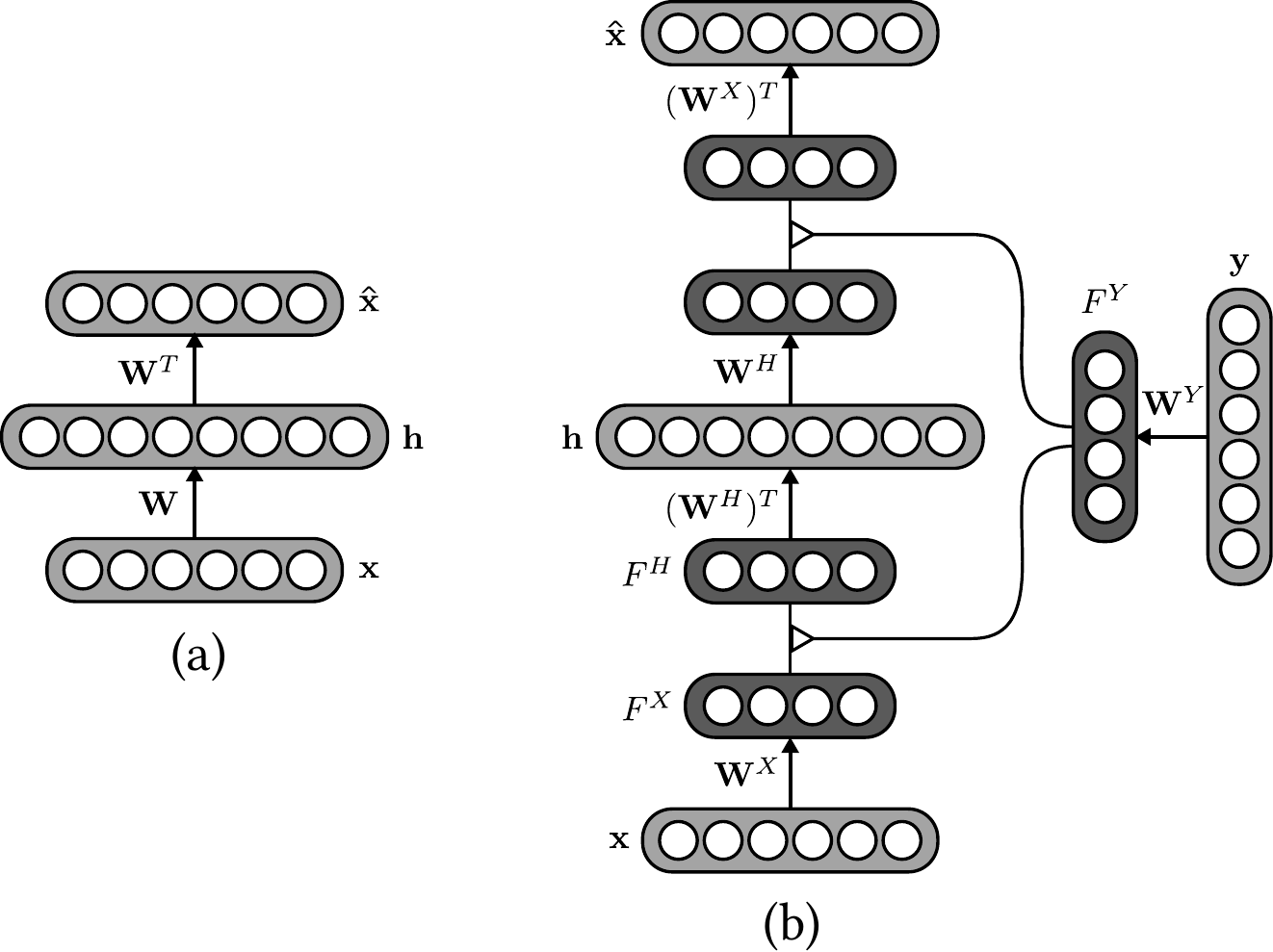}
  \end{center}
  \caption{Illustration of a typical DAE (a) and GAE (b) model each
    with a single hidden layer. Biases have been
  omitted for clarity.}
  \label{fig:archs}
\end{figure}

Gated autoencoders (GAE), also called Relational autoencoders, are an extension
the autoencoder framework which learn relations on input-output
pairs $\x \in \Reals^{n_X}$ given $\y \in \Reals^{n_Y}$ \citep{memisevic2013learning}.
Instead of defining a fixed weight matrix $\W \in \Reals^{n_H \times n_X}$, the GAE
learns a function $w(\y)$ where the model weights are modulated by
the value of the conditioning variable $\y$.

The naive implementation involves constructing a weight tensor $\W \in \Reals^{n_H \times n_X \times n_Y}$
and defining 
\begin{equation}
  \label{eq:gae_w}
  w_{ij}(\y) = \sum_{k=1}^{n_Y} \W_{ijk}y_k
\end{equation}

where the subscripts indicate indexing. Under this model, the encoder
in Equation \ref{eq:ae_enc} becomes
\begin{equation}
  \h = f(\x, \y) = s_H(w(\y) \x + \bias)
  \label{eq:gae_enc}
\end{equation}

However, this requires storing and learning 
$(n_X \times n_Y \times n_H)$  model parameters. In practice, this is infeasible
for all but the smallest problems as the number of weights is cubic in
the number of units (assuming $n_X$, $n_Y$ and $n_H$ are roughly equal). However, we can restrict the interactions
between $\x$, $\y$, and $\h$ by first projecting each onto factors
$F^{X}, F^{Y}, F^{H} \in \Reals^{n_F}$ and allowing only element-wise interaction
between the factors. Instead of a quadratic number of weights needed in the naive method above,
the factored model is parameterized via three weight matrices: 
$\W^X \in \Reals^{n_X \times n_F}, \W^Y \in \Reals^{n_Y \times n_F}$ and $\W^H \in \Reals^{n_F \times n_H} $.
The hidden representation under the factored model is defined as

\begin{equation}
  \h = f(\x, \y) = s_H \left(\left(\W^H\right)^T \left( \W^X \x \otimes \W^Y \y \right) + \bias^H \right)
  \label{eq:gae_fac_enc}
\end{equation}

where $\otimes$ denotes elementwise multiplication and $\bias^H \in \Reals^{n_H}$ is the hidden bias. Notice that the encoder is a function
over both $\x$ and $\y$. Similarly, the decoder function will also be over two input
variables, the choice of which being dependant on which of $\x$ and $\y$ are
to be reconstructed. When learning a conditional model, one of the input variables
will be held fixed. For example, the reconstruction of $\x$ given $\y$ is defined as

\begin{equation}
  \hat \x = g(\h, \y) = s_O \left(\left(\W^X\right)^T \left( \W^H \h \otimes \W^Y \y \right) + \bias^X \right)
  \label{eq:gae_fac_dec}
\end{equation}

where $\bias^H \in \Reals^{n_X}$ is the output bias and $s_O$ is the output activation function.
Equations \ref{eq:gae_fac_enc} and \ref{eq:gae_fac_dec} describe a symmetric model
where the encoder and decoder share the same set of weight matrices. However, this is not 
a hard requirement and various regimes of tied vs untied weights for gated models have been explored
by \citet{alain2013gated}.

Like the traditional autoencoder, the GAE is typically trained
with a denoising criterion where the noise is applied to both $\x$ and $\y$ inputs.
Yet where a traditional autoencoder learns features of the input, a GAE
can learn relationships between its two inputs. For example, when trained on pairs
of images where one is translated version of the other, the filters learned by
the input factors resemble phase-shifted Fourier components \citep{memisevic2013learning}.

The loss function is defined in much the same way as the classical autoencoder
described above, using squared error on real-valued inputs and cross entropy
loss where the input values are binary. As all operations are differentiable,
the model can be trained via stochastic gradient descent while making use
of any optimization techniques (e.g.~momentum, adaptive learning
rates, RMSprop, etc.) that have been developed for neural network training.

\section{Gated Autoencoders as class-conditional Generative models}
\label{sec:classcond}
The sampling procedure proposed by \citet{bengio2013generalized} for classical denoising
autoencoders can also be applied to a GAE. Here,
we define the true data distribution as the conditional distribution
$\mathcal{P}(\x|\y)$. Here $\x \in \Reals^{n_X}$ is an input data point and $\y \in \Reals^{n_Y}$ 
is the associated class label in a `one-hot' encoding. Although GAE training typically applies noise to both $\x$ and $\y$, we
will examine the case where the corruption procedure is applied to $\x$ only.
Thus, the corruption distribution is the same as in the DAE
framework, namely the noise procedure draws samples
from $C(\tilde \x | \x)$. By choosing a loss function which
is interpretable as log-likelihood, the GAE
learns the conditional distribution $P_\theta(\x | \tilde \x, \y)$

Like the sampling procedure for DAEs, the Markov chain formed by alternating samples from 
$\x_t \sim P_\theta(\x|\tilde \x_{t-1}, \y)$ and $\tilde \x_{t} \sim C(\tilde \x | \x)$
will generate samples from the true distribution $\mathcal{P}(\x | \y)$.
During training, we can also apply a class-conditional version
of the walkback training algorithm to seek out and squash any spurious
modes of our model.

\citet{bengio2013generalized} provide a proof of the following
theorem:

\begin{thm}
  If $P_\theta(\x|\tilde \x)$ is a consistent estimator of the true conditional
  distribution $\mathcal{P}(\x|\tilde \x)$ and $T_n$ defines an ergodic
  Markov chain, then as the number of examples $n \rightarrow \infty$,
  the asymptotic distribution $\pi_n(\x)$ of the generated samples converges
  to the data-generating distribution $\mathcal{P}(\x)$.
\end{thm}

For the conditional case of GAE the same arguments hold, where each
of the following substitutions are made:
$P_\theta(\x| \tilde \x)$ with $P_\theta(\x| \tilde \x, \y)$, 
$\mathcal{P}(\x | \tilde \x)$ with $\mathcal{P}(\x | \tilde \x, \y)$
and $\mathcal{P}(\x)$ with $\mathcal{P}(\x | \y)$.
The arguments for consistency and ergoticity can also be made in
the same manner as those in \citep{bengio2013generalized}.

Geometrically, this model can be seen as learning a conditional
manifold of the data. Like the DAE, the model learns to correct the
noisy input by pushing it back towards the data manifold. However,
the model makes use of the class labels in order to learn a separate
manifold for each class label. The $F^Y$ factors, via their multiplicative
interactions, scale the features learned by the $F^X$ factors depending
on the class. It is akin to learning a DAE for each class, yet the gated
model can make use of cross-class structure by sharing weights between
each of the class-specific models.

When training on images of hand written digits, for example, 
the `tail' of a 4, 7, or 9 may all make use of the same feature learned by
one of the $F^X$ factors. As such, the $F^Y$ factor for these classes
would have a relatively large value for this shared feature. However,
this tail feature may not be useful for reconstruction of a 3, 8 or 6.
For these classes, the $F^Y$ values that correspond to the tail 
feature can learn to down-weight the importance of this tail feature.
Under this interpretation, the $F^Y$ factors are a means
of weighting each of the $F^X$ factors dependent on the conditioning
class label.

\section{Experiments}
We demonstrate the generative properties of the class-conditional GAE model
on two datasets: binarized MNIST \citep{lecun1998mnist} and the Toronto Face
Database (TFD) \citep{susskind2010toronto}. The MNIST database of handwritten
digits consists of 60,000 training examples and 10,000 test examples. Pixel
intensities of each 28 $\times$ 28 image have been scaled to $[0, 1]$ and thresholded
at $0.5$, such that each pixel takes on a binary value from $\{0,
1\}$. 

For the MNIST dataset, a GAE was trained with 1024 factors and 512 hidden units.
The hidden units used a rectified linear (ReLU) activation function, while the visible activation
was the logistic (i.e.~sigmoid) function. Using a cross entropy loss, 
the model was trained for 200 epochs via mini-batch gradient descent with a batch size of 100. The
initial learning rate of 0.25 was decreased multiplicatively at each epoch
by a factor of 0.995. For optimization, the model was trained using Nesterov's
accelerated gradient \citep{sutskever2013importance}
 with a parameter of 0.9.
Salt and pepper noise was applied during training where each pixel
has a 0.5 probability of corruption. Training followed the walkback
training procedure, with the training loss averaged over 5 reconstructions
from the Markov chain starting at the training example.

\begin{figure}
  \begin{center}
    \includegraphics{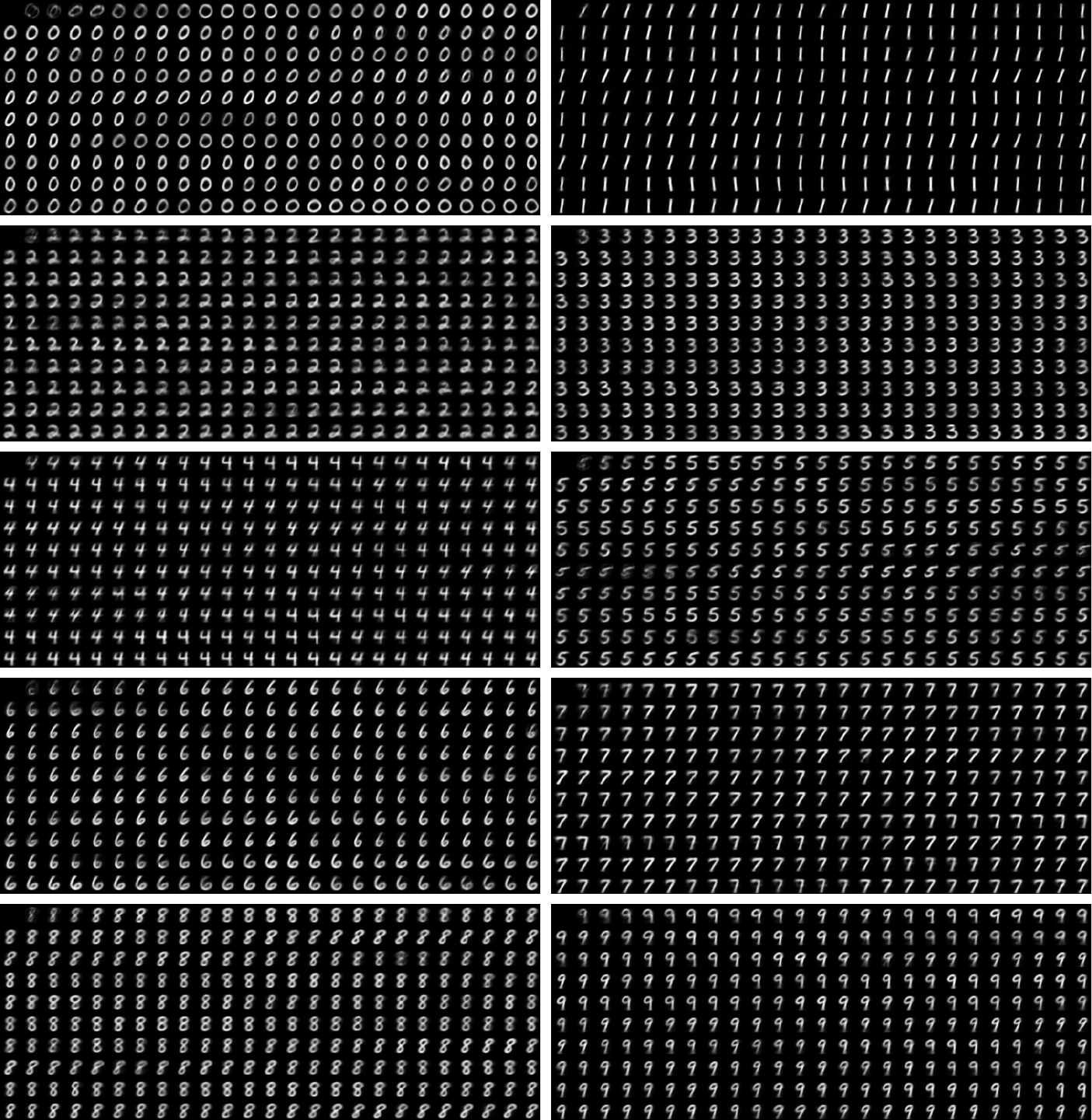}
  \end{center}
  \caption{Consecutive samples generated by the Markov chain learned by the
  class-conditional GAE model trained on binarized MNIST.}
  \label{fig:mnist_samples}
\end{figure}

Figure \ref{fig:mnist_samples} shows 250 consecutive samples generated
while conditioning
on each class label. Note that the `samples' depicted show the expected
value of each pixel for each step of the Markov chain defined by the sampling procedure.
Each chain was initialized to a vector of zeros and at each step, the
image was corrupted with 0.5 salt-and-pepper noise. The model begins to generate
convincing samples after only a few steps of the chain. Also notice that
the chain exhibits mixing between modes in the data distribution. For example,
the samples from the `2' class contains both twos with a loop and a cusp.
Finally, notice that the samples contain very few spurious examples -- i.e.
most samples resemble recognizable digits.

The TFD consists of 4,178 expression-labeled images of human faces. Each image
is 48 $\times$ 48 pixels with 256 integer value levels of gray intensity.
The intensity values were linearly scaled to the range $[0, 1]$. Unlike
the MNIST experiment, the intensity values were not binarized.

A GAE with 512 factors and 1024 ReLU hidden units and sigmoid outputs was
trained on squared-error loss for 500 epochs. Again, we used mini-batch
gradient descent, however, a mini-batch size of 50 was used  with a learning rate
of 1.0 annealed by a factor of 0.995
at each epoch. Training followed 5 walkback steps with a Nesterov momentum
of 0.9. The model was trained and sampled using salt-and-pepper noise with
probability of 0.5.

\begin{figure}
  \begin{center}
    \includegraphics{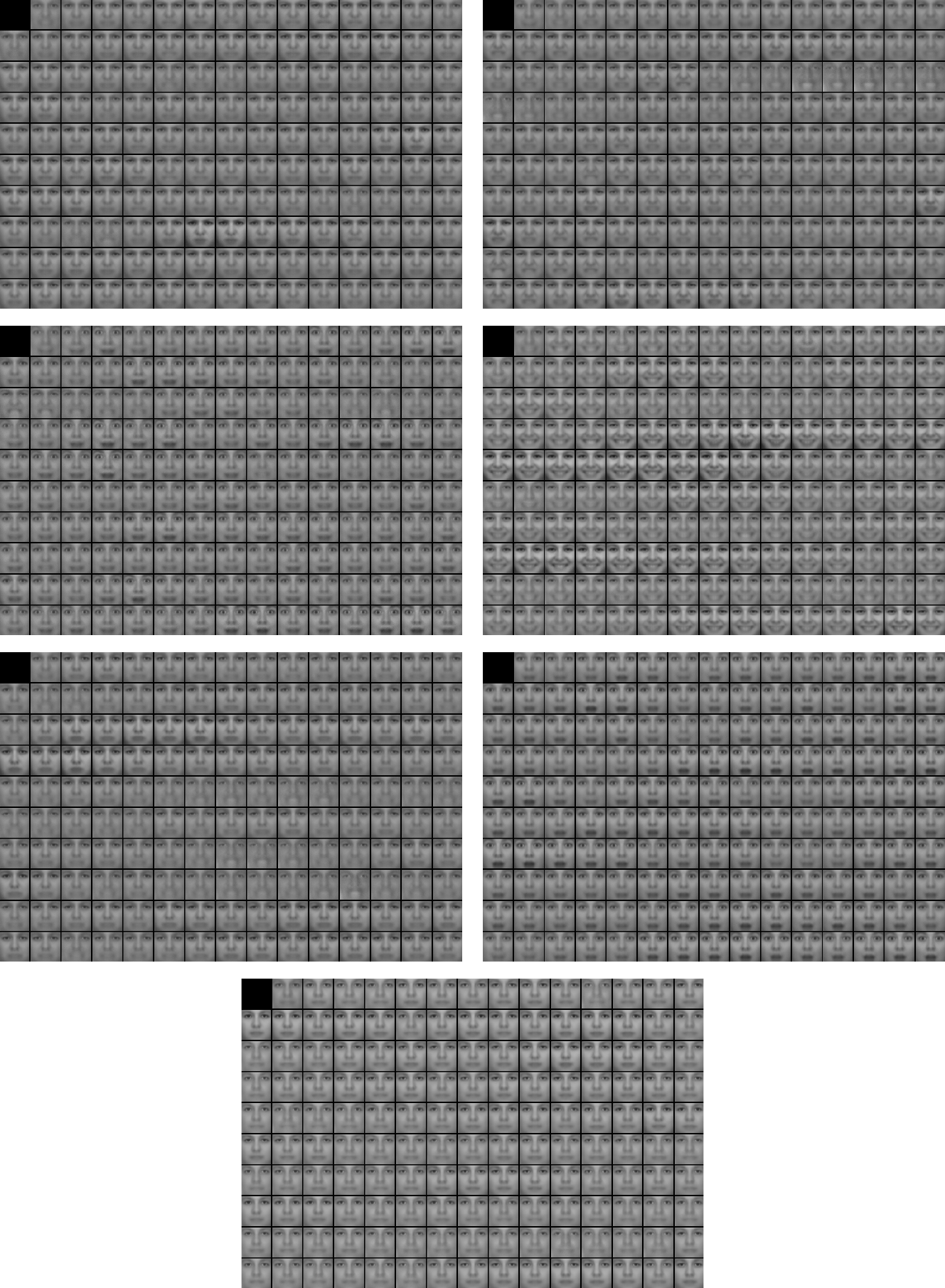}
  \end{center}
  \caption{Consecutive samples generated by the Markov chain learned by the
  class-conditional GAE model trained on TFD images. From left to right,
  top to bottom, the classes are: anger, disgust, fear, happiness, sadness,
  surprise, neutral. 
    }
  \label{fig:tfd_samples}
\end{figure}

When training on MNIST, each walkback reconstruction was then 
sampled using the sigmoid activation as the probabilities of a
factorized Bernoulli distribution. Since the TFD faces were not binarized, the sigmoid
activations were used instead. Figure \ref{fig:tfd_samples} shows
150 consecutive samples for each class label, each starting
at the zero vector.

Notice that there is less variation in the TFD samples than those from
the MNIST model. This is likely due to the relative small size of the TFD
training set (4,178 TFD training cases vs. 60,000 MNIST training cases).
The TFD dataset also provides 112,234 unlabeled examples and it may be possible to pretrain
the $F^X$ factor weights on this unlabeled data. However, these
experiments are beyond the scope of the current analysis.

\section{Conclusion}
The class-conditional GAE can be interpreted as a generative model
in much the same ways as the DAE. In fact, the GAE is akin to learning
a separate DAE model for each class, but with significant weight sharing
between the models. In this light, the gating acts as a means
of modulating the model's weights depending on the class label.
As such, the theoretical and practical consideration that have been
applied to DAE's as generative models can also be applied to 
gated models.
Future work will apply these techniques to richer conditional distributions,
such as the task of image tagging as explored by \citet{mirza2014conditional}.

\bibliography{ccdae}{}

\begin{thebibliography}{21}
\providecommand{\natexlab}[1]{#1}
\providecommand{\url}[1]{\texttt{#1}}
\expandafter\ifx\csname urlstyle\endcsname\relax
  \providecommand{\doi}[1]{doi: #1}\else
  \providecommand{\doi}{doi: \begingroup \urlstyle{rm}\Url}\fi

\bibitem[Alain \& Olivier(2013)Alain and Olivier]{alain2013gated}
Alain, Droniou and Olivier, Sigaud.
\newblock Gated autoencoders with tied input weights.
\newblock In \emph{Proceedings of The 30th International Conference on Machine
  Learning}, pp.\  154--162, 2013.

\bibitem[Bengio \& Thibodeau-Laufer(2013)Bengio and
  Thibodeau-Laufer]{bengio2013deep}
Bengio, Yoshua and Thibodeau-Laufer, {\'E}ric.
\newblock Deep generative stochastic networks trainable by backprop.
\newblock \emph{arXiv preprint arXiv:1306.1091}, 2013.

\bibitem[Bengio et~al.(2013)Bengio, Yao, Alain, and
  Vincent]{bengio2013generalized}
Bengio, Yoshua, Yao, Li, Alain, Guillaume, and Vincent, Pascal.
\newblock Generalized denoising auto-encoders as generative models.
\newblock In \emph{Advances in Neural Information Processing Systems}, pp.\
  899--907, 2013.

\bibitem[Dahl et~al.(2010)Dahl, Marc'Aurelio~Ranzato, Mohamed, and
  Hinton]{dahl2010phone}
Dahl, George~E, Marc'Aurelio~Ranzato, Abdel-rahman~Mohamed, Mohamed,
  Abdel-rahman, and Hinton, Geoffrey~E.
\newblock Phone recognition with the mean-covariance restricted boltzmann
  machine.
\newblock In \emph{NIPS}, pp.\  469--477, 2010.

\bibitem[Deng et~al.(2010)Deng, Seltzer, Yu, Acero, Mohamed, and
  Hinton]{deng2010binary}
Deng, Li, Seltzer, Michael~L, Yu, Dong, Acero, Alex, Mohamed, Abdel-Rahman, and
  Hinton, Geoffrey~E.
\newblock Binary coding of speech spectrograms using a deep auto-encoder.
\newblock In \emph{Interspeech}, pp.\  1692--1695. Citeseer, 2010.

\bibitem[Goodfellow et~al.(2014{\natexlab{a}})Goodfellow, Pouget-Abadie, Mirza,
  Xu, Warde-Farley, Ozair, Courville, and Bengio]{goodfellow2014generative}
Goodfellow, Ian, Pouget-Abadie, Jean, Mirza, Mehdi, Xu, Bing, Warde-Farley,
  David, Ozair, Sherjil, Courville, Aaron, and Bengio, Yoshua.
\newblock Generative adversarial nets.
\newblock In \emph{Advances in Neural Information Processing Systems}, pp.\
  2672--2680, 2014{\natexlab{a}}.

\bibitem[Goodfellow et~al.(2014{\natexlab{b}})Goodfellow, Bulatov, Ibarz,
  Arnoud, and Shet]{goodfellow2014multi}
Goodfellow, Ian~J, Bulatov, Yaroslav, Ibarz, Julian, Arnoud, Sacha, and Shet,
  Vinay.
\newblock Multi-digit number recognition from street view imagery using deep
  convolutional neural networks.
\newblock In \emph{ICLR}, 2014{\natexlab{b}}.

\bibitem[Kamyshanska \& Memisevic(2013)Kamyshanska and
  Memisevic]{kamyshanska2013autoencoder}
Kamyshanska, Hanna and Memisevic, Roland.
\newblock On autoencoder scoring.
\newblock In \emph{Proceedings of the 30th International Conference on Machine
  Learning (ICML-13)}, pp.\  720--728, 2013.

\bibitem[Kingma \& Welling(2014)Kingma and Welling]{kingma2014auto}
Kingma, Diederik~P and Welling, Max.
\newblock Auto-encoding variational bayes.
\newblock In \emph{Proceedings of the International Conference on Learning
  Representations (ICLR)}, 2014.

\bibitem[Krizhevsky et~al.(2012)Krizhevsky, Sutskever, and
  Hinton]{krizhevsky2012imagenet}
Krizhevsky, Alex, Sutskever, Ilya, and Hinton, Geoffrey~E.
\newblock Imagenet classification with deep convolutional neural networks.
\newblock In \emph{NIPS}, volume~1, pp.\ ~4, 2012.

\bibitem[Larochelle \& Bengio(2008)Larochelle and
  Bengio]{larochelle2008classification}
Larochelle, Hugo and Bengio, Yoshua.
\newblock Classification using discriminative restricted boltzmann machines.
\newblock In \emph{Proceedings of the 25th international conference on Machine
  learning}, pp.\  536--543. ACM, 2008.

\bibitem[LeCun \& Cortes(1998)LeCun and Cortes]{lecun1998mnist}
LeCun, Yann and Cortes, Corinna.
\newblock The mnist database of handwritten digits, 1998.

\bibitem[Memisevic(2013)]{memisevic2013learning}
Memisevic, Roland.
\newblock Learning to relate images.
\newblock \emph{Pattern Analysis and Machine Intelligence, IEEE Transactions
  on}, 35\penalty0 (8):\penalty0 1829--1846, 2013.
\newblock ISSN 0162-8828.
\newblock \doi{10.1109/TPAMI.2013.53}.

\bibitem[Memisevic et~al.(2010)Memisevic, Zach, Pollefeys, and
  Hinton]{memisevic2010gated}
Memisevic, Roland, Zach, Christopher, Pollefeys, Marc, and Hinton, Geoffrey~E.
\newblock Gated softmax classification.
\newblock In Lafferty, J.D., Williams, C.K.I., Shawe-Taylor, J., Zemel, R.S.,
  and Culotta, A. (eds.), \emph{Advances in Neural Information Processing
  Systems 23}, pp.\  1603--1611. Curran Associates, Inc., 2010.
\newblock URL
  \url{http://papers.nips.cc/paper/3895-gated-softmax-classification.pdf}.

\bibitem[Mirza \& Osindero(2014)Mirza and Osindero]{mirza2014conditional}
Mirza, Mehdi and Osindero, Simon.
\newblock Conditional generative adversarial nets.
\newblock In \emph{NIPS 2014 Workshop on Deep Learning}, 2014.

\bibitem[Ozair et~al.(2013)Ozair, Yao, and Bengio]{ozair2013multimodal}
Ozair, Sherjil, Yao, Li, and Bengio, Yoshua.
\newblock Multimodal transitions for generative stochastic networks.
\newblock \emph{arXiv preprint arXiv:1312.5578}, 2013.

\bibitem[Susskind et~al.(2010)Susskind, Anderson, and
  Hinton]{susskind2010toronto}
Susskind, Josh~M, Anderson, Adam~K, and Hinton, Geoffrey~E.
\newblock The toronto face database.
\newblock \emph{Department of Computer Science, University of Toronto, Toronto,
  ON, Canada, Tech. Rep}, 2010.

\bibitem[Sutskever et~al.(2013)Sutskever, Martens, Dahl, and
  Hinton]{sutskever2013importance}
Sutskever, Ilya, Martens, James, Dahl, George, and Hinton, Geoffrey.
\newblock On the importance of initialization and momentum in deep learning.
\newblock In \emph{Proceedings of the 30th International Conference on Machine
  Learning (ICML-13)}, pp.\  1139--1147, 2013.

\bibitem[Szegedy et~al.(2014)Szegedy, Liu, Jia, Sermanet, Reed, Anguelov,
  Erhan, Vanhoucke, and Rabinovich]{szegedy2014going}
Szegedy, Christian, Liu, Wei, Jia, Yangqing, Sermanet, Pierre, Reed, Scott,
  Anguelov, Dragomir, Erhan, Dumitru, Vanhoucke, Vincent, and Rabinovich,
  Andrew.
\newblock Going deeper with convolutions.
\newblock \emph{arXiv preprint arXiv:1409.4842}, 2014.

\bibitem[Vincent et~al.(2008)Vincent, Larochelle, Bengio, and
  Manzagol]{vincent2008extracting}
Vincent, Pascal, Larochelle, Hugo, Bengio, Yoshua, and Manzagol,
  Pierre-Antoine.
\newblock Extracting and composing robust features with denoising autoencoders.
\newblock In \emph{Proceedings of the 25th international conference on Machine
  learning}, pp.\  1096--1103. ACM, 2008.

\bibitem[Vincent et~al.(2010)Vincent, Larochelle, Lajoie, Bengio, and
  Manzagol]{vincent2010stacked}
Vincent, Pascal, Larochelle, Hugo, Lajoie, Isabelle, Bengio, Yoshua, and
  Manzagol, Pierre-Antoine.
\newblock Stacked denoising autoencoders: Learning useful representations in a
  deep network with a local denoising criterion.
\newblock \emph{The Journal of Machine Learning Research}, 9999:\penalty0
  3371--3408, 2010.

\end{thebibliography}
\bibliographystyle{iclr2015}
\end{document}